\documentclass[runningheads]{llncs}

\usepackage[year=2024]{eccv}

\usepackage{eccvabbrv}

\usepackage{graphicx}
\usepackage{booktabs}

\usepackage[accsupp]{axessibility}  

\usepackage{hyperref}

\usepackage{orcidlink}

\usepackage{url}

\usepackage[utf8]{inputenc} 
\usepackage[T1]{fontenc}    
\usepackage{hyperref}       
\usepackage{url}            
\usepackage{booktabs}       
\usepackage{amsfonts}       
\usepackage{nicefrac}       
\usepackage{microtype}      

\usepackage{subcaption}
\usepackage[ruled]{algorithm2e} 
\usepackage{epsfig} 
 \usepackage{amsmath}
\usepackage{amssymb} 

\usepackage{epstopdf}

\def\etal{\emph{et al.}}

\def\ie{\emph{i.e.}}

\newcommand{\bs}{\boldsymbol}

\begin{document}

\title{Handling The Non-Smooth Challenge in Tensor SVD: A Multi-Objective Tensor Recovery Framework   } 
 
\titlerunning{Handling The Non-Smooth Challenge in Tensor SVD}

\author{Jingjing Zheng\inst{1}\orcidlink{0000-0003-1955-5308} \and
Wanglong Lu\inst{2}\orcidlink{0000-0001-7956-4084}  \and
Wenzhe Wang\inst{3} \and Yankai Cao\inst{4}\thanks{Corresponding Author} \orcidlink{0000-0001-9014-2552} \and Xiaoqin Zhang\inst{5}\orcidlink{0000-0003-0958-7285} \and Xianta Jiang\inst{2} \orcidlink{0000-0002-3219-1871}}

\authorrunning{J.~Zheng et al.}

\institute{Department of Mathematics, The University of British Columbia, British Columbia, Canada  
\and Department of Computer Science, Memorial University of Newfoundland, Newfoundland and Labrador, Canada \and Information Institute,
Qingdao Preschool Education College, Shandong, China \and Department of Chemical and Biological Engineering, The University of British Columbia, British Columbia, Canada  
\and College of Computer Science and Technology, Zhejiang University of Technology, Zhejiang, China}

\maketitle

\begin{abstract}
Recently, numerous tensor singular value decomposition (t-SVD)-based tensor recovery methods have shown promise in processing visual data, such as color images and videos. However, these methods often suffer from severe performance degradation when confronted with tensor data exhibiting non-smooth changes. It has been commonly observed in real-world scenarios but ignored by the traditional t-SVD-based methods. In this work, we introduce a novel tensor recovery model with a learnable tensor nuclear norm to address such challenge. We develop a new optimization algorithm named the Alternating Proximal Multiplier Method (APMM) to iteratively solve the proposed tensor completion model. Theoretical analysis demonstrates the convergence of the proposed APMM to the Karush–Kuhn–Tucker (KKT) point of the optimization problem. In addition, we propose a multi-objective tensor recovery framework based on APMM to efficiently explore the correlations of tensor data across its various dimensions, providing a new perspective on extending the t-SVD-based method to higher-order tensor cases. Numerical experiments demonstrated the effectiveness of the proposed method in tensor completion.


\end{abstract}

\section{Introduction}\label{section:Introduction} 
In recent years,   lots of tensor methods have been proposed  to better analyze the low-rankness in massive high-dimensional tensor data, including color images, hyperspectral images, and videos \cite{kajo2019self, lu2019tensor,zhang2014novel, lu2019low, lu2018exact,madathil2018twist}, as the traditional matrix methods \cite{candes2010matrix,candes2009exact,candes2011robust,chandrasekaran2009sparse,xu2012robust,wright2009robust} fail on handling the tensor data. Depending on different adopted low-rank prior,  these tensor methods can be categorized as: (1) CP (Canonical Polyadic) Decomposition-based methods \cite{hitchcock1927expression,hitchcock1928multiple,kolda2009tensor}, (2)  Tucker Decomposition-based methods \cite{tucker1963implications,gandy2011tensor,liu2013tensor}, and (3) t-SVD-based methods   \cite{lu2019low,lu2018exact,ZHENG2020170,Qin2022Low}.

 Recently,  the t-SVD-based methods   have  gained increasing attention and achieved great success in the applications of visual data processing, such as data denoising \cite{lu2019tensor}, image and video inpainting \cite{zhang2014novel, lu2019low, lu2018exact},  and background modeling and initialization \cite{  lu2019tensor,kajo2018svd}. In these t-SVD-based methods, a fixed invertible transform, such as Discrete Fourier Transform (DFT) \cite{lu2018exact} and  Discrete Cosine Transform (DCT) \cite{lu2019low}, is applied to a  tensor data along its certain dimension. The low-rankness prior of each frontal slice of the transformed tensor has been used to explore global low-rankness of tensor data. However, the approach of performing invertible transform and analyzing the low rankness  along   a specific dimension of tensor data  poses the following  challenges, namely: (i) non-smooth challenge, and (ii) the lack of an effective way to generalize  t-SVD-based methods directly to higher-order tensor cases.
 
  \begin{figure}[tb]
  \centering
    \begin{subfigure}{0.3\linewidth} \includegraphics[width=1.4in]{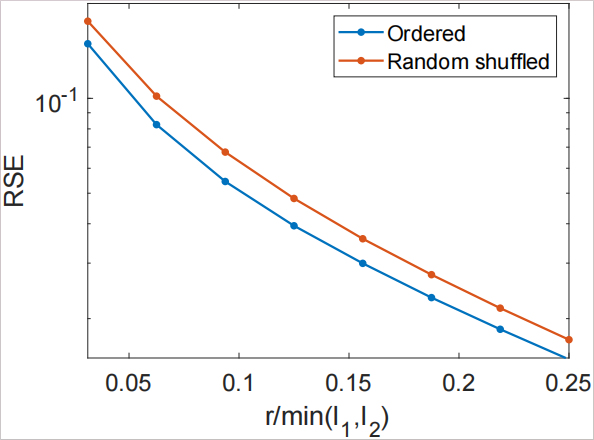}
    \caption{Comparison of DFT-based t-SVD on yale dataset in ordered and random shuffled.}
    \label{fig1-a}
  \end{subfigure}
  \hfill
   \begin{subfigure}{0.3\linewidth} \includegraphics[width=1.4in]{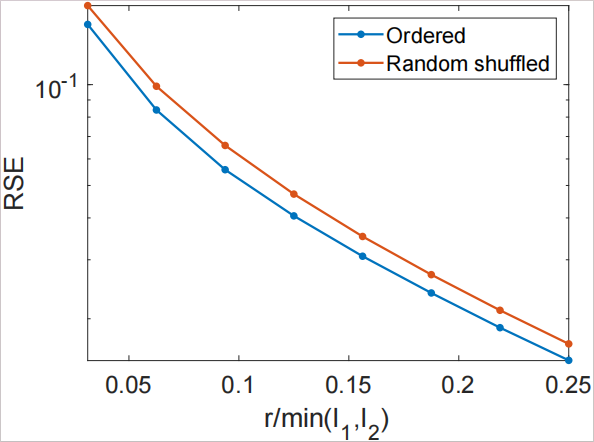}
    \caption{Comparison of DCT-based t-SVD on yale dataset in ordered and random shuffled.}
    \label{fig1-b}
  \end{subfigure}
  \hfill
  \begin{subfigure}{0.3\linewidth} \includegraphics[width=1.2in]{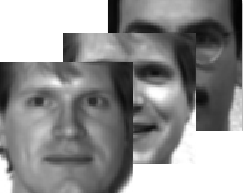}
    \caption{Ordered images.}
    \label{fig1-c}
  \end{subfigure}
  \hfill
  \begin{subfigure}{0.3\linewidth} \includegraphics[width=1.4in]{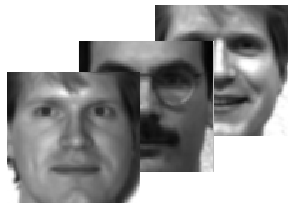}
    \caption{Disordered images for classification task.}
    \label{fig1-d}
  \end{subfigure}
    \hfill
  \begin{subfigure}{0.3\linewidth}  \includegraphics[width=1.4in]{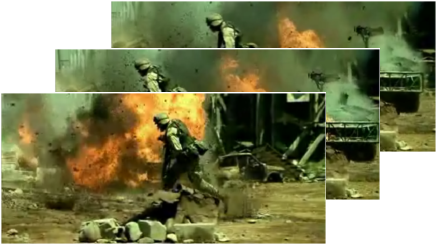}
      \caption{Video with rapidly changing frames}\label{fig1-e}
  \end{subfigure}
    \hfill
  \begin{subfigure}{0.3\linewidth}
\includegraphics[width=1.3in]{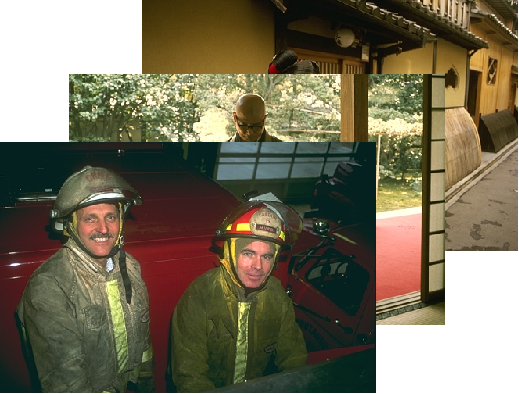}
    \caption{Image sequence with different scenes.}
    \label{fig1-f}
  \end{subfigure}
  \caption{Illustration to challenges of t-SVD-based methods in real world   scenarios.}
  \label{fig1}
\end{figure}

Regarding the non-smooth challenge, Fig.~\ref{fig1} illustrates two different real-world scenarios: disordered images sequence (Fig.~\ref{fig1} (a)-(d)) and  tensor data with non-smooth changes (Fig.~\ref{fig1} (e)-(f)). These  commonly encountered scenarios pose a significant challenge for t-SVD-based tensor analysis methods. Taken an  example  illustrated in Fig.~\ref{fig1} (a)-(d), the disordered images sequence often affect  the tensor recovery performance significantly. This phenomenon is referred as to tensor slices permutation variability (SPV) \cite{Zheng_Zhang_Wang_Jiang_2022}, \ie, interchanging the frontal slice of the tensor will affect the t-SVD results. Since the sequence of samples is often disordered prior to classification, this phenomenon is frequently observed within classification tasks.  Although  Zheng \etal have proposed an effective solution for handling such issue by solving a Minimum Hamiltonian circle problem for the case of DFT \cite{Zheng_Zhang_Wang_Jiang_2022}, a general solution is still lacking.  
In addition, the recovery performance of the t-SVD-based methods is also susceptible to  non-smooth of the tensor data itself \cite{kong2021tensor}. These data include videos with rapidly changing content between frames and  tensor data obtained by concatenating images with different scenes as illustrated in Fig.~\ref{fig1} (e) and (f), respectively. These non-smooth challenges  arise because a fixed invertible transform, such as DFT or DCT, is applied to the tensor along certain dimensions, making the t-SVD sensitive to disorder and non-smooth changes in tensor slices. 

To address the mentioned second challenge, a common solution for handling high-order tensors is to utilize tensor unfolding operators \cite{ZHENG2020170,Qin2022Low}. For example, in \cite{ZHENG2020170}, the Weighted Sum of Tensor Nuclear Norm of all mode-$(k_1,k_2)$ unfolding tensors (WSTNN) has been proposed to  investigate   correlations along different modes of higher order tensors. However, the consideration of weighted summation in WSTNN results in a challenging setting requiring $h(h-1)/2$  weight parameters. Therefore, there is an urgent need for more effective methods to address this issue.

This study  aims to address the above two   challenges, and our contributions are as follows. 
\begin{itemize}  
\item We proposed a new tensor recovery model with a learnable  tensor nuclear norm by introducing a set of unitary matrices  to effectively address SPV and non-smoothness issues in the traditional t-SVD-based methods, thereby allowing our model to harness the inherent data characteristics. 

\item We presented a novel optimization algorithm named the Alternating Proximal Multiplier Method (APMM) to solve the proposed  tensor recovery model effectively,  along with a corresponding convergence analysis in theory.   
\item We are the first  in the literature to propose   a   multi-objective tensor recovery with learned tensor nuclear norms for effectively  exploring the low-rankness of tensor data across its various dimensions, without the need for introducing numerous tensor variables and weights as in traditional weighted sum-based methods \cite{ZHENG2020170,zhang2022tensor}. The experimental results demonstrates the superior performance in tensor completion than other methods. For instance,  
the proposed framework has achieved a 3.5 dB improvement in color vodeo inpaiting!

\end{itemize}

		 \begin{table}[tb]
	\centering
	\caption{Notations}\label{notation}
	\vspace{0.1cm}
	\renewcommand\arraystretch{1.5} 
		\resizebox{\textwidth}{!}{ 
			\begin{tabular}{|c|c|c|c|}
				\hline  
				Notations                          & Descriptions                & Notations                                      & Descriptions                       \\ \hline
	   $ a $, $b$, $c$, $\cdots$   &  scalars         &     $ \bs{a}$, $ \bs{b}$, $ \bs{c}$, $\cdots$          & vectors        \\       
 \hline       $\bs{A}$, $\bs{B}$, $\bs{C}$, $\cdots$ &  matrices &	$ \bs{\mathcal{A}} $, $ \bs{\mathcal{B}} $, $ \bs{\mathcal{C}} $, $\cdots$                        &   tensors   \\ \hline 
     $\mathbb{A}$, $\mathbb{B}$, $\mathbb{C}$, $\cdots$             & sets 	   &        $\mathcal{A}$, $\mathcal{B}$, $\mathcal{C}$, $\cdots$	   &           operators                            \\ \hline
            $[1:n]$ & $\{1,2,\cdots,n\}$
   &	                  $\bs{A}^T$   &      transpose of $\bs{A}$     \\\hline  
            $\bs{\mathcal{I}}$ &identity tensor
   &	                      $ \bs{0}$     &  null tensor    \\\hline 
          $ [\bs{\mathcal{A}}]_{i_1,i_2,\cdots,i_h}$  &     $(i_1, i_2, \cdots, i_h)$-th element of $\bs{\mathcal{A}}$   &    $\|\bs{\mathcal{A}}\|_0$     & the number of non-zero elements of $\bs{\mathcal{A}}$        \\ \hline  
                      $\|\bs{\mathcal{A}}\|_1$     &  $\|\bs{\mathcal{A}}\|_1=\sum_{i_1,i_2,\cdots,i_h}|[\bs{\mathcal{A}}]_{i_1,i_2,\cdots,i_h}|$  &
          $\bs{\mathcal{A}}^{T_{k_1,\cdots,k_h}}$  &   rotation
 of  $\bs{\mathcal{A}} \in \mathbb{R}^{I_1\times\cdots\times I_h}$ such that $\bs{\mathcal{A}}^{T_{k_1,\cdots,k_h}}\in \mathbb{R}^{I_{k_1}\times\cdots\times I_{k_h}}$    \\\hline        $ [\bs{\mathcal{A}}^{T_{k_1,\cdots,k_h}}]_{:,:, i_{k_3}, \cdots, i_{k_h}}$  &    slice along the $(k_1 ,k_2)$-th mode     &         $\bs{\mathcal{A}}_{(k)}$   (or $[\bs{\mathcal{A}}]_{(n)}$) & Mode-n Unfolding of $\bs{\mathcal{A}}$
             \\\hline  
              $\times_n$ & Mode-n product  &    $\mathbb{A}^c$ &complementary set of $\mathbb{A}$   \\ \hline
       \end{tabular}} 
 
\end{table}

\section{Multi-Objective Tensor Recovery with   Learnable Tensor Nuclear Norms}

 Before introducing our problem and the proposed methods,  we summarize notations in Table \ref{notation} that will be used later.

\subsection{Tensor Completion Problem} 
 Considering tensor  data $\bs{\mathcal{M}} \in \mathbb{R}^{I_1 \times  I_2 \times \cdots  \times I_h}$ may with missing elements, \cite{lu2019low,Qin2022Low} propose  the following t-SVD-based tensor completion model:
 \begin{equation}\label{TC} \min_{\bs{\mathcal{X}}}  ~\mathrm{rank}_{[k_1,k_2]}(\bs{\mathcal{X}})\quad 
 s.t.~ \Psi_{\mathbb{I}}(\bs{\mathcal{M}})=\Psi_{\mathbb{I}}(\bs{\mathcal{X}}), 
 \end{equation}
where $\Psi_{\mathbb{I}}$ is a linear project operator on the support set $\mathbb{I}$ such that 
  $$
[\Psi_{\mathbb{I}}(\bs{\mathcal{M}})]_{i_1,i_2,\cdots,i_h} = \left\{ \begin{array}{l} [\bs{\mathcal{M}}]_{i_1,i_2,\cdots,i_h} , ~~~~~~~  \text{if} ~ (i_1,i_2,\cdots,i_h)\in \mathbb{I};\\
	0, ~~ \text{if} ~ (i_1,i_2,\cdots,i_h)\notin \mathbb{I},
\end{array} \right.$$
and $i_k \in [1:I_k]$ for $k=1,2,\cdots,h$. 
The   tensor rank function in \eqref{TC}  operates on the assumption that an $h$-order tensor data $\bs{\mathcal{M}} \in \mathbb{R}^{I_1 \times  I_2 \times \cdots  \times I_h}$ in the real world can be  decomposed as
\begin{equation}\label{D11} 
\bs{\mathcal{M}}=\bs{\mathcal{Z}}  \times_{k_3} \bs{\hat{U}}^T_{k_3} \cdots \times_{k_h} \bs{\hat{U}}^T_{k_h},
\end{equation} 
where $\{k_3, k_4, \cdots, k_h\} \subset [1:h]$, $\{\bs{\hat{U}}_{k_n}\}^h_{n=3}$  are a set of given  invertible transforms,   and  \begin{align}\label{slicerank} \mathrm{rank}_{[k_1,k_2]}(\bs{\mathcal{M}})&=\max_{i_{k_n} \in [1:I_{k_n}] \text{~for~} n=3,4,\cdots,h}\mathrm{rank}\Big([\bs{\mathcal{Z}}^{T_{k_1,\cdots,k_h}}]_{:,:,i_{k_3}, \cdots,i_{k_h}}\Big) \notag \\
&<< \min(I_{k_1}, I_{k_2})\end{align} 
 for certain $(k_1, k_2)$ satisfying $1 \leq k_1<k_2\leq h$. If taken $\{\bs{\hat{U}}_{k_n}\}_{n=3}^h$  as Discrete   Fourier  Matrices, $\mathrm{rank}_{[k_1,k_2]}(\bs{\mathcal{M}})$ is referred as to tensor tubal rank of $\bs{\mathcal{M}}$ \cite{lu2018exact}.


\subsection{The Proposed Tensor Completion with A Learnable Tensor Rank Function} 

 From the definition of  
$\mathrm{rank}_{[k_1,k_2]}(\bs{\mathcal{M}})$, we can see that it aims to examine the slice-wise tensor rank along the $(k_1, k_2)$-th mode of  transformed data $\bs{\mathcal{Z}}$. It allows  \eqref{TC} to investigate the low-rankness of different features in tensor data corresponding to different frequencies separately and jointly.

However, when  $\bs{\mathcal{M}}$ exhibits harsh changes caused by disordered tensor slices along a certain mode $k_n$, all information in $\bs{\mathcal{M}}$ tends to collapse into  the high-frequency slices in $\bs{\mathcal{Z}}$.  To handle such situation, we introduce  a learnable  permutation matrix $\bs{P}\in \mathbb{R}^{I_{k_n}\times I_{k_n} }$ to \eqref{D11}:   
   $$\bs{\mathcal{M}} \times_{k_n} \bs{P} = \bs{\mathcal{Z}} \times_{k_3} \bs{\hat{U}}^T_{k_3} \cdots \times_{k_n} (\bs{\hat{U}}_{k_n}\bs{P})^T \times_{k_{n+1}} \cdots \times_{k_h} \bs{\hat{U}}^T_{k_h},$$
   and therefore we can  get the following model: \begin{align}\label{Hard}\min_{\bs{\mathcal{X}},\bs{P}_{k_n}}& \mathrm{rank}_{[k_1,k_2]}(\bs{\mathcal{X}}\times_{k_n} \bs{P}_{k_n})\notag\\
 s.t.&~ \Psi_{\mathbb{I}}(\bs{\mathcal{M}})=\Psi_{\mathbb{I}}(\bs{\mathcal{X}}), \notag \\
  &~\sum_{i=1}^{I_{k_n}}[\bs{P}_{k_n}]_{i,j}=1, ~\sum_{j=1}^{I_{k_n}}[\bs{P}_{k_n}]_{i,j}=1  \text{~for~}[\bs{P}_{k_n}]_{i,j}\in\{0,1\}.    
 \end{align} 
   This incorporation is aimed at addressing the slice permutation property in tensor completion methods, thus facilitating a more effective exploration of the low-rank property in $\bs{\mathcal{M}}$. 
   
  Unfortunately,  solving \eqref{Hard} is challenging due to the constraints of $\sum_{i=1}^{I_{k_n}}[\bs{P}_{k_n}]_{i,j}=1$ and  $\sum_{j=1}^{I_{k_n}}[\bs{P}_{k_n}]_{i,j}=1$ for $  [\bs{P}_{k_n}]_{i,j}\in\{0,1\}$.  Therefore,  we opt to use  a set of learnable unitary matrices $\{\bs{U}_{k_n}\}_{n=s+1}^h$  instead,   proposing  the following tensor completion model for   given $\{k_{s+1},k_{s+2},\cdots, k_{h}\}$: \begin{align}\label{TCSU}\min_{\bs{\mathcal{X}},\bs{U}_{k_n}(n=s+1,\cdots,h)}& \mathrm{rank}_{[k_1,k_2]}(\bs{\mathcal{X}}\times_{k_{s+1}} \bs{U}_{k_{s+1}} \cdots \times_{k_h} \bs{U}_{k_h})\notag\\
 s.t.&~ \Psi_{\mathbb{I}}(\bs{\mathcal{M}})=\Psi_{\mathbb{I}}(\bs{\mathcal{X}}),~\bs{U}^T_{k_n}\bs{U}_{k_n}=\bs{I}(n=s+1,\cdots,h),   
 \end{align}
 where $\bs{\mathcal{X}}$ is a low-rank estimation of the true tensor data $\bs{\mathcal{M}}$. Besides simplifying the optimization process of \eqref{Hard}, the introduction of the learnable unitary matrices enables the model to extract features  for  better studying the low-rankness in tensor data and  to handle other scenarios where the initially provided transforms $\{\bs{\hat{U}}_{k_n}\}_{n=3}^h$ are inadequate, such as videos with irregularly changing content and image sequences with different scenes.
Hence, we introduce our learnable tensor rank as $$\mathrm{rank}_{[k_1,k_2],\tilde{\mathcal{U}}}(\bs{\mathcal{X}})=\mathrm{rank}_{[k_1,k_2]}(\tilde{\mathcal{U}}(\bs{\mathcal{X}})),$$ where  $\tilde{\mathcal{U}}(\bs{\mathcal{X}})=\bs{\mathcal{X}}\times_{k_{s+1}} \bs{U}_{k_{s+1}} \cdots \times_{k_h} \bs{U}_{k_h}$.


\subsection{Approximation to The Proposed Tensor Completion by Using A Learnable Tensor Nuclear Norm (TC-SL)}  
 Since    the function $\mathrm{rank}_{[k_1, k_2],\tilde{\mathcal{U}}}(\cdot)$  
is discrete, it often leads to the NP-hard problem.
From the result given in \cite{lu2019low, lu2018exact}, we know that, when $h=3$ and $\bs{\hat{U}}_3$ is orthogonal,  $\|\cdot\|^{[1,2]}_{*}$ is the tightest convex envelope of $\mathrm{rank}_{[1, 2]}(\cdot)$ on the set $\{\bs{\mathcal{A}}|\|\bs{\mathcal{A}}\|^{[1,2]}_{2}\leq 1\}$, where   the definitions of $\|\cdot\|^{[k_1,k_2]}_{2}$ and $\|\cdot\|^{[k_1,k_2]}_{*}$ are given in Definition \ref{hnorms}. This conclusion can be easily extended to our case, and we obtain the Property \ref{p1}. 
\begin{definition}\label{hnorms}  
  For an h-order  tensor $\bs{\mathcal{A}} \in \mathbb{R}^{I_{1} \times I_{2}  \times \cdots \times I_{h}}$,  $\|\bs{\mathcal{A}}\|^{[k_1,k_2]}_{*}$ and $ \|\bs{\mathcal{A}}\|^{[k_1,k_2]}_{2} $ are defined as  $$\|\bs{\mathcal{A}}\|^{[k_1,k_2]}_{*} =\sum_{i_{k_3}, i_{k_4}, \cdots, i_{k_h}}\|[\bs{\mathcal{A}}^{T_{k_1,\cdots,k_h}}\times_{k_3} \bs{\hat{U}}_{k_3} \cdots \times_{k_h} \bs{\hat{U}}_{k_h}]_{:,:, i_{k_3}, \cdots, i_{k_h}}\|_*$$ and $$\|\bs{\mathcal{A}}\|^{[k_1,k_2]}_{2} =\max_{i_{k_3}, i_{k_4}, \cdots, i_{k_h}}\|[\bs{\mathcal{A}}^{T_{k_1,\cdots,k_h}}\times_{k_3} \bs{\hat{U}}_{k_3} \cdots \times_{k_h} \bs{\hat{U}}_{k_h}]_{:,:, i_{k_3},  \cdots, i_{k_h}}\|_2,$$ respectively.
\end{definition}
\begin{property}\label{p1}   
  For an h-order  tensor $\bs{\mathcal{A}} \in \mathbb{R}^{I_{1} \times I_{2}  \times \cdots \times I_{h}}$, let us define  $\|\bs{\mathcal{A}}\|^{[k_1,k_2]}_{*, \tilde{\mathcal{U}}}$ and $ \|\bs{\mathcal{A}}\|^{[k_1,k_2]}_{2, \tilde{\mathcal{U}}} $ are defined as  $ \|\bs{\mathcal{A}}\|^{[k_1,k_2]}_{*, \tilde{\mathcal{U}}} = \| \tilde{\mathcal{U}}(\bs{\mathcal{A}})\|^{[k_1,k_2]}_{*} $  and $ \|\bs{\mathcal{A}}\|^{[k_1,k_2]}_{2, \tilde{\mathcal{U}}} =\|\tilde{\mathcal{U}}(\bs{\mathcal{A}})\|^{[k_1,k_2]}_{2}$, respectively.$\|\cdot\|^{[k_1,k_2]}_{*, \tilde{\mathcal{U}}}$ is the dual norm of    tensor $\|\cdot\|^{[k_1,k_2]}_{2,\tilde{\mathcal{U}}}$ norm, and  $\|\cdot\|^{[k_1,k_2]}_{*, \tilde{\mathcal{U}}}$ is the tightest convex envelope of $\mathrm{rank}_{[k_1, k_2], \tilde{\mathcal{U}}}(\cdot)$ on the set $\{\bs{\mathcal{A}}|\|\bs{\mathcal{A}}\|^{[k_1,k_2]}_{2, \tilde{\mathcal{U}}} \leq 1\}$.  \end{property} 

Therefore, we derive an approximation to the proposed model \eqref{TCSU} by utilizing a learnable tensor nuclear norm  based on   the prior assumption of slice-wise low-rankness in the  transformed data  (TC-SL): 
 \begin{equation}\label{TCSU} \min_{\bs{\mathcal{X}},\tilde{\mathcal{U}}} \|\bs{\mathcal{X}}\|^{[k_1,k_2]}_{*, \tilde{\mathcal{U}}}  
 \qquad s.t.~ \Psi_{\mathbb{I}}(\bs{\mathcal{M}})=\Psi_{\mathbb{I}}(\bs{\mathcal{X}}).  
 \end{equation}
Let $\mathcal{\hat{U}}$ be the  invertible transform operator learned by    \eqref{TCSU}.  The exactly recovery  of TC-SL with given $\mathcal{\hat{U}}$ is guaranteed from the current studies in the exactly recovery of the t-SVD-based tensor completion     \cite{lu2019low,lu2018exact}. 

\subsection{The Proposed Multi-Objective  Tensor Completion to Learn The Cross-Dimensional Low-Rankness}

From  \eqref{TCSU}, we observe that the definition of TC-SL  depends on the choice of  $k_1$ and $k_2$, and it considers different kinds of low-rankness  in the tensor data by  adjusting $(k_1, k_2)$. However, considering only one mode may result in the loss of correlation information across the remaining modes. To address this issue, we give the following multi-objective model with  learnable tensor nuclear norms for tensor completion (MOTC): 
 \begin{equation}\label{MOTC}  \min_{\bs{\mathcal{X}}, \tilde{\mathcal{U}}_{(k_1,k_2)}} \Big[\|\bs{\mathcal{X}}\|^{[k_1,k_2]}_{*, \tilde{\mathcal{U}}_{(k_1,k_2)}}\Big]_{1 \leq k_1<k_2\leq h}  \qquad s.t.~\Psi_{\mathbb{I}}(\bs{\mathcal{M}})=\Psi_{\mathbb{I}}(\bs{\mathcal{X}}),
 \end{equation}
 where $\tilde{\mathcal{U}}_{(k_1,k_2)}(\bs{\mathcal{X}})= \bs{\mathcal{X}}\times_{k_{s+1}} \bs{U}^{(k_1,k_2)}_{k_{s+1}} \cdots \times_{k_h} \bs{U}^{(k_1,k_2)}_{k_h}$ and  $(\bs{U}_{k_n}^{(k_1,k_2)})^T\bs{U}_{k_n}^{(k_1,k_2)}=\bs{I}$ for $n=1+s,\cdots,h$ and $1 \leq k_1<k_2\leq h$. In MOTC, the multiple objective functions $\|\bs{\mathcal{X}}\|^{[k_1,k_2]}_{*, \tilde{\mathcal{U}}_{(k_1,k_2)}} (1 \leq k_1<k_2\leq h)$ are utilized to examine the low-rankness of tensor data from its various dimensions. 

\section{Optimization  Algorithm }  
In this section, we  provide a detailed discussion for the optimizing    \eqref{TCSU} and \eqref{MOTC}.
\subsection{Alternating Proximal Multiplier Method (APMM) for  TC-SL}


To solve the  problem \eqref{TCSU}, we  introduce   auxiliary variables  $\bs{\mathcal{E}} \in \mathbb{E}= \{\bs{\mathcal{E}}|\Psi_{\mathbb{I}}(\bs{\mathcal{E}})=\textbf{0} \}$ and $\bs{\mathcal{Z}}$  such that $\bs{\mathcal{X}}=\bs{\mathcal{Z}} \times_{k_{s+1}} \bs{U}^T_{k_{s+1}} \cdots \times_{k_h} \bs{U}^T_{k_h}$. Therefore, we turn to solve the following equivalence problem:
       \begin{equation}\label{TCBTQ_auxi2} 
\min_{\bs{\mathcal{Z}},\bs{U}^T_{k_n}\bs{U}_{k_n}=\bs{I}(n=s+1,\cdots,h)} \|\bs{\mathcal{Z}}\|^{[k_1,k_2]}_{*} \qquad
s.t. ~ \Psi_{\mathbb{I}}(\bs{\mathcal{M}})= \bs{\mathcal{Z}} \times_{k_{s+1}} \bs{U}^T_{k_{s+1}} \cdots \times_{k_h} \bs{U}^T_{k_h}+\bs{\mathcal{E}}.
 \end{equation}
 The Augmented Lagrangian function of \eqref{TCBTQ_auxi2} is formulated as 
\begin{align}\label{TCBTQ_auxi_La}
&\mathcal{L}(\bs{\mathcal{Z}},\{\bs{U}_{k_n}\}_{n=s+1}^h,\bs{\mathcal{E}},\bs{\mathcal{Y}},\mu)=\|\bs{\mathcal{Z}}\|^{[k_1,k_2]}_{*}+\langle\Psi_{\mathbb{I}}(\bs{\mathcal{M}})-\bs{\mathcal{Z}}\times_{k_{s+1}} \bs{U}^T_{k_{s+1}} \cdots \times_{k_h} \bs{U}^T_{k_h} \notag \\
&-\bs{\mathcal{E}}, \bs{\mathcal{Y}}\rangle
+\dfrac{\mu}{2}\|\Psi_{\mathbb{I}}(\bs{\mathcal{M}})-\bs{\mathcal{Z}}\times_{k_{s+1}} \bs{U}^T_{k_{s+1}} \cdots \times_{k_h} \bs{U}^T_{k_h}-\bs{\mathcal{E}}\|_F^2,
\end{align}
  where $\bs{\mathcal{Y}}$ is Lagrange multiplier, and $ \mu $ is a positive scalar. We solve  \eqref{TCBTQ_auxi2} iteratively by combining the proximal algorithm with the Alternating Direction Method of Multipliers (APMM) that
is given in the Algorithm \ref{Alg2}.
 We detail the solutions for  solving $\bs{\mathcal{Z}}^{(t+1)}$, $\bs{U}_{k_{n}}^{(t+1)}$, and $\bs{\mathcal{E}}^{(t+1)}$  as follows.\\ 
 \begin{itemize} \item[$\bullet$]\textbf{Calculate $\bs{\mathcal{Z}}^{(t+1)}$:}
\begin{align} \bs{\mathcal{Z}}^{(t+1)}=\mathop{\arg\min}_{\bs{\mathcal{Z}}}&\dfrac{1}{2}\|\dfrac{\mu^{(t)}\hat{\bs{\mathcal{P}}}\times_{k_{s+1}} \bs{U}^{(t)}_{k_{s+1}} \cdots \times_{k_h} \bs{U}^{(t)}_{k_h}+\eta^{(t)}\bs{\mathcal{Z}}^{(t)}}{\mu^{(t)}+\eta^{(t)}}-\bs{\mathcal{Z}}\|_F^2 \notag \\
&\qquad\qquad\qquad\qquad\qquad\qquad\qquad+\dfrac{1}{\mu^{(t)}+\eta^{(t)}}\|\bs{\mathcal{Z}}\|^{[k_1,k_2]}_{*},\notag
\end{align} 
where  $\hat{\bs{\mathcal{P}}}=\Psi_{\mathbb{I}}(\bs{\mathcal{M}})-\bs{\mathcal{E}}^{(t)}+\dfrac{1}{\mu^{(t)}}\bs{\mathcal{Y}}^{(t)}$. It can be solved by the tensor singular value thresholding operation with parameter $\dfrac{1}{\mu^{(t)}+\eta^{(t)}}$ \cite{lu2019tensor}.\\
\item[$\bullet$] \textbf{Calculate $\bs{U}_{k_{n}}^{(t+1)}$:}
Let $\bs{\mathcal{A}}=\bs{\mathcal{Z}}^{(t+1)}\times_{k_h} \bs{U}^{(t)T}_{k_h} \times_{k_h-1} \cdots \times_{{k_{n}}+1} \bs{U}^{(t)T}_{{k_{n}}+1}$ and  
$\bs{\mathcal{B}}=\bs{\hat{\mathcal{P}}}\times_{s+1} \bs{U}^{(t+1)}_{k_{s+1}}\cdots \times_{{k_{n}}-1} \bs{U}^{(t+1)}_{{k_{n}}-1}$, and 
we have 
$$\bs{U}_{k_{n}}^{(t+1)}=	\mathop{\arg\min}_{\bs{U}^T_{k_{n}}\bs{U}_{k_{n}}=\bs{I}}\|\bs{U}_{k_{n}}[\sqrt{\mu^{(t)}}\bs{\mathcal{B}}_{(k_{n})}, \sqrt{\eta^{(t)}}\bs{I}]-[\sqrt{\mu^{(t)}}\bs{\mathcal{A}}_{(k_{n})}, \sqrt{\eta^{(t)}}\bs{U}^{(t)}_{k_{n}}]\|_F^2.$$
The optimal solution $\bs{U}_{k_{n}}^{(t+1)}$ can be gievn by   $\bs{U}_{k_{n}}^{(t+1)}=\bs{U}\bs{V}^T$from \cite{zou2006sparse}, where  
$\bs{U}$ and $\bs{V}$ can be obtained by SVD of $\mu^{(t)}\bs{\mathcal{A}}_{({k_{n}})}\bs{\mathcal{B}}_{({k_{n}})}^T+ \eta^{(t)}\bs{U}^{(t)}_{k_{n}}$: $\mu^{(t)}\bs{\mathcal{A}}_{({k_{n}})}\bs{\mathcal{B}}_{({k_{n}})}^T+ \eta^{(t)}\bs{U}^{(t)}_{k_{n}}=\bs{U}\bs{\Sigma}\bs{V}^T$.\\
\item[$\bullet$] \textbf{Calculate $\bs{\mathcal{E}}^{(t+1)}$:}  
\begin{align}\label{update_e} \bs{\mathcal{E}}^{(t+1)}=&\mathop{\arg\min}_{\bs{\mathcal{E}}\in \mathbb{E}}\dfrac{\mu^{(t)}}{2}\|\Psi_{ \mathbb{I}}(\bs{\mathcal{M}})-\bs{\mathcal{X}}^{(t+1)}-\bs{\mathcal{E}}+\dfrac{1}{\mu^{(t)}}\bs{\mathcal{Y}}^{(t)}\|_F^2+\dfrac{\eta^{(t)}}{2}\|\bs{\mathcal{E}}-\bs{\mathcal{E}}^{(t)}\|^2_F \notag\\
 =&\Psi_{\mathbb{I}^c}( \dfrac{1}{\mu^{(t)}+\eta^{(t)}}(\mu^{(t)}(\Psi_{\mathbb{I}}(\bs{\mathcal{M}})-\bs{\mathcal{X}}^{(t+1)}+\dfrac{1}{\mu^{(t)}}\bs{\mathcal{Y}}^{(t)})+\eta^{(t)}\bs{\mathcal{E}}^{(t)})). \notag
\end{align}
\end{itemize}
\begin{algorithm}[t]
			\caption{ APMM-based Iterative Solver   to \eqref{TCBTQ_auxi2} }\label{Alg2}  \KwIn{$\Psi_{\mathbb{I}}(\bs{\mathcal{M}})$,   $\{\bs{U}_{k_n}^{(0)}\}_{n=s+1}^h$, $\{\bs{\hat{U}}_{k_n}^{(0)}\}_{n=3}^h$, $\bs{\mathcal{E}}^{(0)}$, $\bs{\mathcal{Y}}^{(0)}$, $t=0$,   $\mu^{(0)}$, $\eta^{(0)}$, $\rho_{\mu}, \rho_{\eta}>1$,   $\bar{\mu}$, and $\bar{\eta}$.} \KwOut{  $\bs{\mathcal{Z}}^{(t+1)}$ ,$\{\bs{U}_{k_n}^{(t+1)}\}_{n=s+1}^h$, and $\bs{\mathcal{X}}^{(t+1)}$.}
   1.~\textbf{While not converge do}\\
		2.\quad $\bs{\mathcal{Z}}^{(t+1)}= \mathop{\arg\min}_{\bs{\mathcal{Z}}}\mathcal{L}(\bs{\mathcal{Z}},\{\bs{U}^{(t)}_{k_n}\}_{n=s+1}^h,\bs{\mathcal{E}}^{(t)},\bs{\mathcal{Y}}^{(t)},\mu^{(t)})+ \dfrac{\eta^{(t)}}{2}\|\bs{\mathcal{Z}}^{(t)}-\bs{\mathcal{Z}}\|^2_F$;
\\ 
          3.\quad Calculate    $\bs{U}_{k_{n}}^{(t+1)}=	\mathop{\arg\min}_{\bs{U}^T_{k_{n}}\bs{U}_{k_{n}}=\bs{I}}\mathcal{L}(\bs{\mathcal{Z}}^{(t+1)},\{\bs{U}^{(t+1)}_{k_n}\}_{n=s+1}^{n-1},$\\
          
          $\bs{U}_{k_{n}},\{\bs{U}^{(t)}_{k_n}\}_{n=n+1}^h,\bs{\mathcal{E}}^{(t)}, \bs{\mathcal{Y}}^{(t)},\mu^{(t)})+\dfrac{\eta^{(t)}}{2}\|\bs{U}^{(t)}_{k_{n}}-\bs{U}_{k_{n}}\|_F^2,~(s+1 \leq n_0 \leq h)$;\\

4.\quad Calculate $\bs{\mathcal{X}}^{(t+1)}$ by $\bs{\mathcal{X}}^{(t+1)}=\bs{\mathcal{Z}}^{(t+1)}\times_{k_h} \bs{U}^{(t+1)T}_{k_h} \times_{k_{h-1}} \cdots \times_{k_{s+1}} \bs{U}^{(t+1)T}_{k_{s+1}}$;\\
          
          5.\quad  $\bs{\mathcal{E}}^{(t+1)}=\mathop{\arg\min}_{\bs{\mathcal{E}}\in \mathbb{E}}\mathcal{L}(\bs{\mathcal{Z}}^{(t+1)},\{\bs{U}^{(t+1)}_{k_n}\}_{n=s+1}^h,\bs{\mathcal{E}},\bs{\mathcal{Y}}^{(t)},\mu^{(t)})+\dfrac{\eta^{(t)}}{2}\|\bs{\mathcal{E}}-\bs{\mathcal{E}}^{(t)}\|^2_F$;
\\
6.\quad Calculate $\bs{\mathcal{Y}}^{(t+1)}$ by 	$\bs{\mathcal{Y}}^{(t+1)}=\mu^{(t)}(\Psi_{\mathbb{I}}(\bs{\mathcal{M}})-\bs{\mathcal{X}}^{(t+1)}-\bs{\mathcal{E}}^{(t+1)})+\bs{\mathcal{Y}}^{(t)}$;
\\
7.\quad Calculate   $\mu^{(t+1)}=\min(\bar{\mu}, \rho_{\mu}\mu^{(t)})$  and $\eta^{(t+1)}=\min(\bar{\eta}, \rho_{\eta}\eta^{(t)})$, respectively;\\ 
          8.\quad Check the convergence condition: $\|\bs{\mathcal{Z}}^{(t+1)}-\bs{\mathcal{Z}}^{(t)}\|_{\infty}<\varepsilon$, $\|\bs{\mathcal{X}}^{(t+1)} -\bs{\mathcal{X}}^{(t)}\|_{\infty}<\varepsilon$, 
          
          ~~~\quad$\| \bs{U}^{(t+1)}_{k_n} -\bs{U}^{(t)}_{k_n}\|_{\infty}<\varepsilon$ for $n=s+1,s+2,\cdots,h$; \\
          9.\quad t=t+1.\\
    10.~\textbf{end while}
		\end{algorithm} 
\subsubsection{Computational Complexity} The most time-consuming steps in the algorithm \ref{Alg2}  are the computations of  $\bs{\mathcal{Z}}$, and $\bs{U}_{k_{n}}$. Since the computational complexity of n-mode product of $\hat{\bs{\mathcal{P}}}\in \mathbb{R}^{I_1\times I_2\times \cdots \times I_h}$ and $\bs{U}_n \in \mathbb{R}^{I_n \times I_n}$ is $\mathcal{O}(I_{n}I_1I_2\cdots I_h)$,  
the   complexity of   the computation of $\bs{\mathcal{Z}}$ is $\mathcal{O}(hI_{(1)}I_1I_2\cdots I_h)$, where $I_{(1)}=\max_k(I_k)$. 
Besides,  
the complexity of   the computation of $\bs{U}_{k_n}$ is 
$\mathcal{O}((h-1)I_{(1)}I_1I_2\cdots I_h+ I_k^2I_1I_2\cdots I_h)$,
therefore  the overall computational complexity of each iteration of Algorithm \ref{Alg2} is $\mathcal{O}\Big((h-s) (h+I_{(1)}-1)I_{(1)}I_1I_2\cdots I_h+ hI_{(1)}I_1I_2\cdots I_h\Big)$.

\subsubsection{Convergence Analysis}\vspace{-0.25cm} 
Although the optimization problem \eqref{TCBTQ_auxi2} is non-convex because of the constraints $\bs{U}^T_{k_n}\bs{U}_{k_n}=\bs{I} (n=s+1, s+2,\cdots,h)$ and the global optimality for  \eqref{TCBTQ_auxi2}  is hardly guaranteed, we can still prove some excellent convergence properties of the Algorithm \ref{Alg2},   as stated in the following theorem.

\begin{theorem}\label{T1}
 For the sequence $\{[\bs{\mathcal{Z}}^{(t)},\{\bs{U}_{k_n}^{(t)}\}_{n=s+1}^h, \bs{\mathcal{E}}^{(t)}, \bs{\mathcal{Y}}^{(t)},\mu^{(t)}]\}$ generated by the proposed algorithm \ref{Alg2},   we have the following properties if $\{\bs{\mathcal{Y}}^{(t)}\}$ is bounded, $\sum_{t=1}^{\infty}(\mu^{(t)})^{-2}\mu^{(t+1)}<+\infty$ and $\lim\limits_{n \longrightarrow \infty} \mu^{(n)}\sum\limits_{t=n}^{\infty} (\eta^{(t)})^{-1/2}=0$.  
\begin{itemize}

\item[(i)] $\lim\limits_{t\longrightarrow \infty}\Psi_{\mathbb{I}}(\bs{\mathcal{M}})-\bs{\mathcal{X}}^{(t)}-\bs{\mathcal{E}}^{(t)}=\bs{0}$, where $\bs{\mathcal{X}}^{(t)}=\bs{\mathcal{Z}}^{(t)}\times_{k_h}(\bs{U}_{k_h}^{(t)})^T\cdots \times_{k_{s+1}}(\bs{U}_{k_{s+1}}^{(t)})^T$. 

\item[(ii)] $\{[\bs{\mathcal{Z}}^{(t)},\{\bs{U}^{(t)}_{k_n}\}_{n=s+1}^h\}, \bs{\mathcal{X}}^{(t)},\bs{\mathcal{E}}^{(t)}]\}$ is bounded.

    \item[(iii)] $\sum_{t=1}^{\infty}\eta^{(t)} \|[\bs{\mathcal{Z}}^{(t)},\{\bs{U}^{(t)}_{k_n}\}_{n=s+1}^h\}, \bs{\mathcal{E}}^{(t)}]-[\bs{\mathcal{Z}}^{(t+1)},\{\bs{U}^{(t+1)}_{k_n}\}_{n=s+1}^h\}, \bs{\mathcal{E}}^{(t+1)}]\|^2_F$ is convergent.    
    Thus, we have  $$\|[\bs{\mathcal{Z}}^{(t)},\{\bs{U}^{(t)}_{k_n}\}_{n=s+1}^h\}, \bs{\mathcal{E}}^{(t)}]-[\bs{\mathcal{Z}}^{(t+1)},\{\bs{U}^{(t+1)}_{k_n}\}_{n=s+1}^h\}, \bs{\mathcal{E}}^{(t+1)}]\|^2_F \leq \mathcal{O}(\dfrac{1}{\eta^{(t)}}).$$ 

   \item[(iv)] $\lim\limits_{t \longrightarrow \infty} \|\bs{\mathcal{Y}}^{(t+1)}-\bs{\mathcal{Y}}^{(t)}\|_F=0$. 
   
      \item[(v)] Let $[\bs{\mathcal{Z}}^{*}, \{\bs{U}^{*}_{k_n}\}_{n=s+1}^h,\bs{\mathcal{E}}^{*},\bs{\mathcal{Y}}^{*}]$ be any limit point of $\{[\bs{\mathcal{Z}}^{(t)}, \{\bs{U}^{(t)}_{k_n}\}_{n=s+1}^h, \bs{\mathcal{E}}^{(t)},$
    
      $ \bs{\mathcal{Y}}^{(t)}]\}$. Then, $[\bs{\mathcal{Z}}^{*}, \{\bs{U}^{*}_{k_n}\}_{n=s+1}^h, \bs{\mathcal{E}}^{*},\bs{\mathcal{Y}}^{*}]$ is  a KKT point  to  \eqref{TCBTQ_auxi2}.
\end{itemize}
	\end{theorem}
Please refer
to the supplementary material of this paper for the proof of the Theorem \ref{T1}.
Theorem \ref{T1} shows that, if $\{\bs{\mathcal{Y}}^{(t)}\}$ is bounded, the sequence ${[\bs{\mathcal{Z}}^{(t)},\{\bs{U}_{k_n}^{(t)}\}_{n=s+1}^h, \bs{\mathcal{E}}^{(t)}]}$ generated by the proposed algorithm \ref{Alg2} is Cauchy convergent, with a convergence rate of at least $\mathcal{O}(\dfrac{1}{\eta^{(t)}})$. Moreover, any accumulation point of the sequence converges to the KKT point of \eqref{TCBTQ_auxi2}.

\subsection{APMM-based Heuristic Method for Solving MOTC}
To solve MOTC, we iteratively update $\bs{U}^{(k_1,k_2)}_{k_n} (n=1+s,\cdots,h, ~1 \leq k_1<k_2\leq h)$ and $\bs{\mathcal{X}}$ in \eqref{MOTC}  by solving  the following optimization problems: 
\begin{equation}\label{TCSU_sub} 
  \hat{\mathcal{U}}_{(k_1,k_2)} 
=\mathop{\arg\min}_{\tilde{\mathcal{U}}_{(k_1,k_2)}}  \|\bs{\hat{\mathcal{X}}}\|^{[k_1,k_2]}_{*, \tilde{\mathcal{U}}_{(k_1,k_2)}}
 \end{equation} 
 and 
 \begin{equation}\label{MOTC_sim} \bs{\hat{\mathcal{X}}}=\mathop{\arg\min}_{\bs{\mathcal{X}}} \Big[\|\bs{\mathcal{X}} \|^{(k_1,k_2)}_{*, \hat{\mathcal{U}}_{(k_1,k_2)}}\Big]_{1 \leq k_1<k_2\leq h}
 s.t.~ \Psi_{\mathbb{I}}(\bs{\mathcal{M}})=\Psi_{\mathbb{I}}(\bs{\mathcal{X}}), 
 \end{equation}
 respectively. 
We  use the proposed APMM for solving \eqref{TCSU_sub}, and the non-dominated sorting genetic algorithm (NSGA-II) \cite{996017} presented in Algorithm  \ref{NSGA}  for solving  \eqref{MOTC_sim}. 
These two steps are used for the learning of    the tensor nuclear norm functions and  the low rank estimation, respectively.  
It is worth noting that both updating   the $  \hat{\mathcal{U}}_{(k_1,k_2)}$ for each $(k_1,k_2)$ and   evaluating the individuals in Algorithm  \ref{NSGA} can be proceed in parallel, which can accelerate the whole optimization for MOTC.

	\begin{algorithm}[t]
			\caption{ NSGA-II-based  framework for solving \eqref{MOTC_sim} }\label{NSGA}  \KwIn{$\Psi_{\mathbb{I}}(\bs{\mathcal{M}})$, $\{\hat{\mathcal{U}}_{(k_1,k_2)}\}_{1\leq k_1 < k_2 \leq h}$,  $\mathrm{IterMax}$, and $t=0$.} \KwOut{ $\bs{\hat{\mathcal{X}}}$}
   1.\quad \textbf{Initialize:}   
   population $\mathbb{P}^{(0)}$\\ 
		2.\quad  Evaluate   $\Big[\|\bs{\mathcal{X}} \|^{[k_1,k_2]}_{*, \hat{\mathcal{U}}_{(k_1,k_2)}}\Big]_{1 \leq k_1<k_2\leq h}$ for each $\bs{\mathcal{X}} \in \mathbb{P}^{(t)}$;
\\ 
          3.\quad Sorting individuals by their non-domination ranks and crowding distance,\\  ~~\quad 
          and obtaining the first front $\mathbb{F}_1$; \\

4.\quad Selecting parents and applying crossover and mutation to create offspring;\\
          
          5.\quad Obtaining the next generation by truncating  the sorted individuals  \\ 
          \quad \quad (including 
 offspring population) and update the first front $\mathbb{F}_1$;  \\
6.\quad $t=t+1$;
\\
7.\quad Repeat steps 2-6 until $t=\mathrm{IterMax}$; \\
8.\quad $\bs{\hat{\mathcal{X}}}$ = average sum of the individuals that belong to $\mathbb{F}_1$.   
		\end{algorithm} 
 

\section{Experimental Results \label{ER}}   
	In this section,   we compared the TC-SL  and MOTC\footnote{The code of our method is available at \url{https://github.com/jzheng20/MOTC.git}.}  with several state-of-the-art methods, including TNN-DCT \cite{lu2019low}, TNN-DFT \cite{lu2018exact}, SNN \cite{liu2013tensor}, KBR \cite{xie2017kronecker},  WSTNN \cite{ZHENG2020170}, and HTNN-DCT \cite{Qin2022Low},  in the context of color video inpainting and images  inpainting. Below are brief explanations for all eight methods: 
 \begin{itemize}
     \item  \textbf{TNN-DCT and TNN-DCT} are
  three-order tensor completion methods;
   \item   \textbf{HTNN-DCT and TC-SL}   are higher-order tensor completion methods, but they consider low-rankness along only one dimension of the tensor;
   \item  \textbf{SNN, KBR,   WSTNN, and MOTC} are higher-order tensor completion methods  that have considered low-rankness across various dimensions of the tensor. SNN and KBR are Tucker decomposition-based methods.
 \end{itemize}
 For the three-order tensor methods, the  data tensors   were constructed by the mode-(1,2) unfolding tensor, \ie, $[\Psi_{\mathbb{I}}(\bs{\mathcal{M}})]_{(1,2)} \in \mathbb{R}^{I_1\times I_2 \times I_3I_4}$.  Here,  $I_1 \times I_2$ represents the size of each image (frame), $I_3=3$ is RGB channel number of each image, and $I_4$ is the number of images (frames). 
 For a fair comparison, we implemented the compared methods using the code provided by the respective authors in our experimental environment. 
 
 We used the Peak Signal-To-Noise Ratio (PSNR) to evaluate the performance of different methods in tensor completion.   To ensure the reliability of our experiments, we conducted each experiment five times and reported the average results as the final outcomes.   The best results for each case are shown in bold.   All experiments were implemented using Matlab R2022b.

\begin{table}[tb]
\centering
\caption{Comparing the PSNR results by different methods on \textit{BSD} at different sampling rates $p$.}
\resizebox{\columnwidth}{!}{%
\begin{tabular}{cccclcccc}
\hline
\multicolumn{1}{c|}{Sampling Rate $p$} &
  TNN-DCT   &
  TNN-DFT   &
  SNN&
  KBR  &
  WSTNN   &
  HTNN-DCT   &
  TC-SL &
 MOTC \\ \hline
\multicolumn{1}{c|}{0.3} &
  23.25 &
  23.21 &
  21.86&
  25.45 &
  25.75 &
  25.21 &
  26.32 &
  \bf{27.53} \\
\multicolumn{1}{c|}{0.5} &
  27.25 &
  27.20 &
  25.50&
  31.57 &
  31.07 &
  30.72 &
  31.55 & \bf{33.25}
    \\
\multicolumn{1}{c|}{0.7} &
  32.04 &
  31.95 &
  29.84&
  38.81 &
  37.11 &
  38.22 &
  38.33 & \bf{39.97}
    \\ \hline 
    \multicolumn{1}{c|}{Average}& 27.51 & 27.45 & 25.73 & 31.94 & 31.31 & 31.38 & 32.06 & \textbf{33.58}  
    \\ \hline 
\end{tabular}%
    }\label{BSD}
\end{table} 

\begin{figure}[tb]
	\centering 
	\includegraphics[width=4.75in]{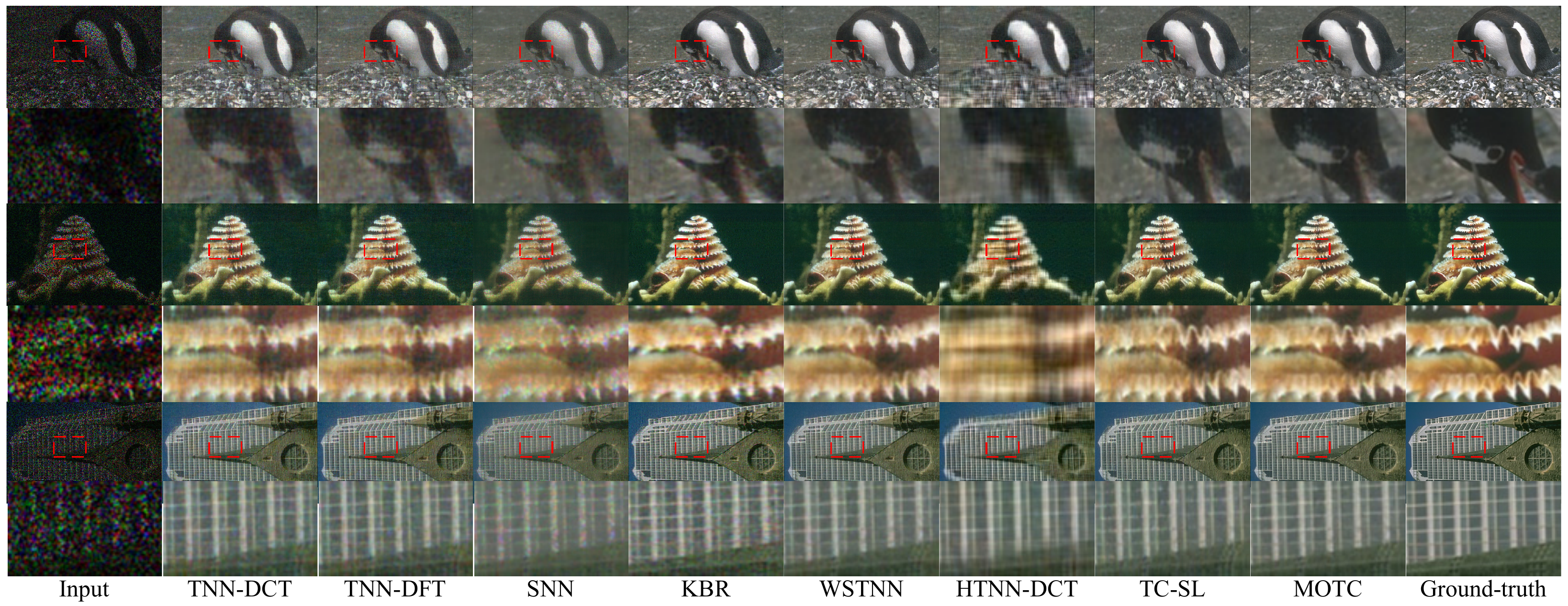} 
	\caption{ Examples of images inpainting by different methods on the BSD dataset with sampling rate $p=0.3$. \textbf{Best viewed in $\times 2$ sized color pdf file.} }\label{fig:fig_bsd_inpainting}
\end{figure}  





\subsection{Images Inpainting}  
\subsubsection{Image  Sequences With Various Scenes}
In this subsection, we evaluated different tensor completion methods on   \textit{Berkeley
Segmentation Dataset} (\textit{BSD})\footnote{\url{https://www2.eecs.berkeley.edu/Research/Projects/CS/vision/bsds/}} \cite{martin2001a}, which includes color  images    with different scenes. Following the experiment setting in \cite{lu2019tensor},  we randomly selected 50 color images for testing. 



Table \ref{BSD} presents the PSNR values achieved by each method. From the table, most of the higher tensor methods (KBR, WSTNN, HTNN-DCT, TC-SL, and MOTC) consistently outperform the three-order tensor methods (TNN-DCT and TNN-DFT) across all cases. It indicates that high-order tensor methods are more effective in accurately capturing the low-rank structure in four-order tensor data.   Additionally,  both TC-SL and MOTC, the proposed methods, exhibit the best performance across all cases, surpassing other methods by approximately 1.5 dB for  $p \in \{0.3, 0.5\}$. Visual examples of images inpainting by different methods at a sampling rate $p=0.3$ are presented in Fig.~\ref{fig:fig_bsd_inpainting}. As observed in the figure, the visual results obtained from WSTNN, TC-SL, and MOTC demonstrate superior reconstruction quality and preservation of finer details compared to other methods. \vspace{-10pt}   

\subsubsection{ Image Sequences With Random Shuffling}

\begin{figure}[tb]
	\centering 
	\includegraphics[width=4.75in]{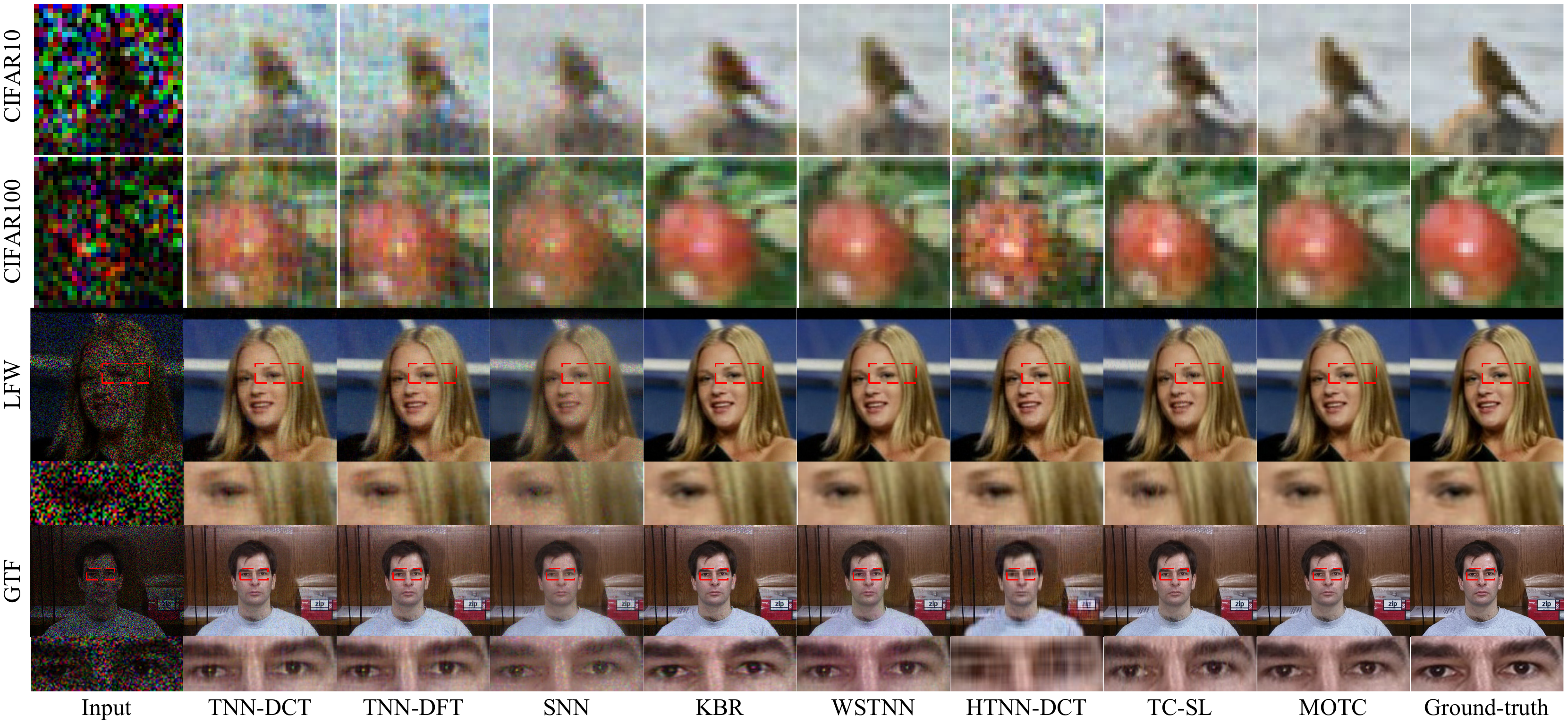} 
	\caption{ Examples of images inpainting  by different methods  on three dataset with sampling rate  $p=0.3$. \textbf{Best viewed in $\times 2$ sized color pdf file.} }\label{fig:fig_image_inpainting}
\end{figure}  
\begin{table}[tb]
\centering
\caption{Comparing the PSNR results by different methods   at sampling rates $p=0.3$.}
\resizebox{\columnwidth}{!}{%
\begin{tabular}{cccclcccc}
\hline 
Data &
  TNN-DCT   &
  TNN-DFT   & 
  SNN &
  KBR  &
   
  WSTNN   &
  HTNN-DCT   &
  TC-SL &
  MOTC \\ \hline
CIFAR10 &19.59 &19.54 & 20.71  & 24.08 &   25.02 & 22.12 &24.63    &  \textbf{26.46} \\
CIFAR100 &19.39 &19.31 & 20.70 &24.15&    24.76  & 21.71& 24.50 & \textbf{26.16}   \\
LFW  &27.49  & 27.45  & 22.48 & 33.47 &  34.33 & 30.15 & 31.57 & \textbf{35.67} \\ 
GTF & 25.46  &25.41 &23.54 & 25.83&    26.66&22.11  &  32.10  & \textbf{33.56} \\\hline  
  Average  &22.98  &  22.92  & 21.85 &26.88 & 27.69   & 24.02  &  28.20 &  \textbf{30.46}  \\ 
  \hline  
\end{tabular}%
}\label{clasi}
\end{table}

In this subsection, we evaluated different tensor completion methods using four image classification datasets: \textit{CIFAR10}\footnote{\url{https://www.cs.toronto.edu/~kriz/cifar.html}}~\cite{krizhevsky2009learning}, \textit{CIFAR100}~\cite{krizhevsky2009learning}, \textit{Labeled Faces in the Wild (LFW)}\footnote{\url{http://vis-www.cs.umass.edu/lfw/}},  and \textit{Georgia Tech Face database (GTF)}\footnote{\url{http://www.anefian.com/research/face_reco.htm}}. Due to constraints in computing resources, we utilized subsets of the datasets. Specifically, for CIFAR10 and CIFAR100, we sampled the first 50 images and 5 images for each category, respectively, resulting in two subsets of 500 images. For LFW, we selected the first 50 classes, and for GTF,  we selected the first five classes.  Random shuffling was applied to the image sequences before testing. All  PSNR results are provided in Table \ref{clasi}.

From the results, both proposed methods (TC-SL and MOTC) achieved the best performance on average. Upon comparing HTNN-DCT and TC-SL, it is observed that the average performance of the proposed TC-SL surpasses HTNN-DCT by more than 4 dB. It indicates that our method, incorporating a learnable tensor norm, effectively handles SPV in t-SVD-based methods. Among all methods,    MOTC  achieves the best performance across all cases. Particularly noteworthy is the average PSNR result obtained by MOTC, outperforming other methods by more than 2.5 dB on average. Additionally, as observed in Fig. \ref{fig:fig_image_inpainting}, the visual results obtained from MOTC exhibit  smoother transitions and better restoration of intricate textures. All results highlights the effectiveness of the proposed framework. \vspace{-10pt} 



  


\subsubsection{Results Analysis}

These results suggest that the proposed   MOTC is able to exploit the correlations of tensor data along different dimensions effectively and can more accurately exploit the low-rank structure of four-order tensor data (the one with non-continue change) than other methods. The superiority of TC-SL  and MOTC over other methods can be attributed to the introduction of both the given transforms 
from smooth priors and learnable unitary matrices, which enable it to better handle the non-smooth in tensor data caused by random shuffling image sequences (such as \textit{CIFAR10}, \textit{CIFAR100}, \textit{LFW}, and \textit{GTF}) or the concatenation of different scene images (such as \textit{BSD}) and capture the underlying low-rank structures in the tensor data more effectively.\vspace{-10pt}

 \begin{figure}[tb]
  \centering
  \caption{Comparing PSNR by different methods on the 50 video segments at a sampling rate $p=0.3$.}
\includegraphics[width=4.8in]{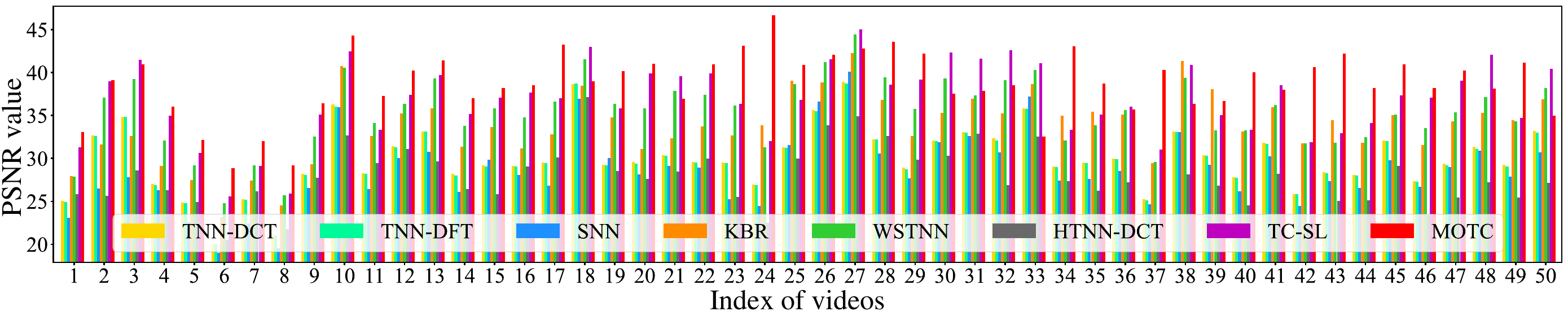}\label{video_bar}
\end{figure}
\begin{table}[tb]
\centering
\caption{Comparing the average PSNR by different methods on the 50 video segments at a sampling rate $p=0.3$.} 
\resizebox{\textwidth}{!}{%
\begin{tabular}{c|cccccccc}
\hline
Video                                   & TNN-DCT  & TNN-DFT  & SNN  & KBR   &     WSTNN & HTNN-DCT  & TC-SL & MOTC\\ \hline
Average                                 & 30.01  & 29.91   & 28.59 & 33.76 &  35.28 & 27.71    & 36.82 &  \textbf{38.74} \\ \hline
\end{tabular}%
}\label{videoe}
\end{table}
  \begin{figure}[tb]
	\centering 
	\includegraphics[width=4.75in]{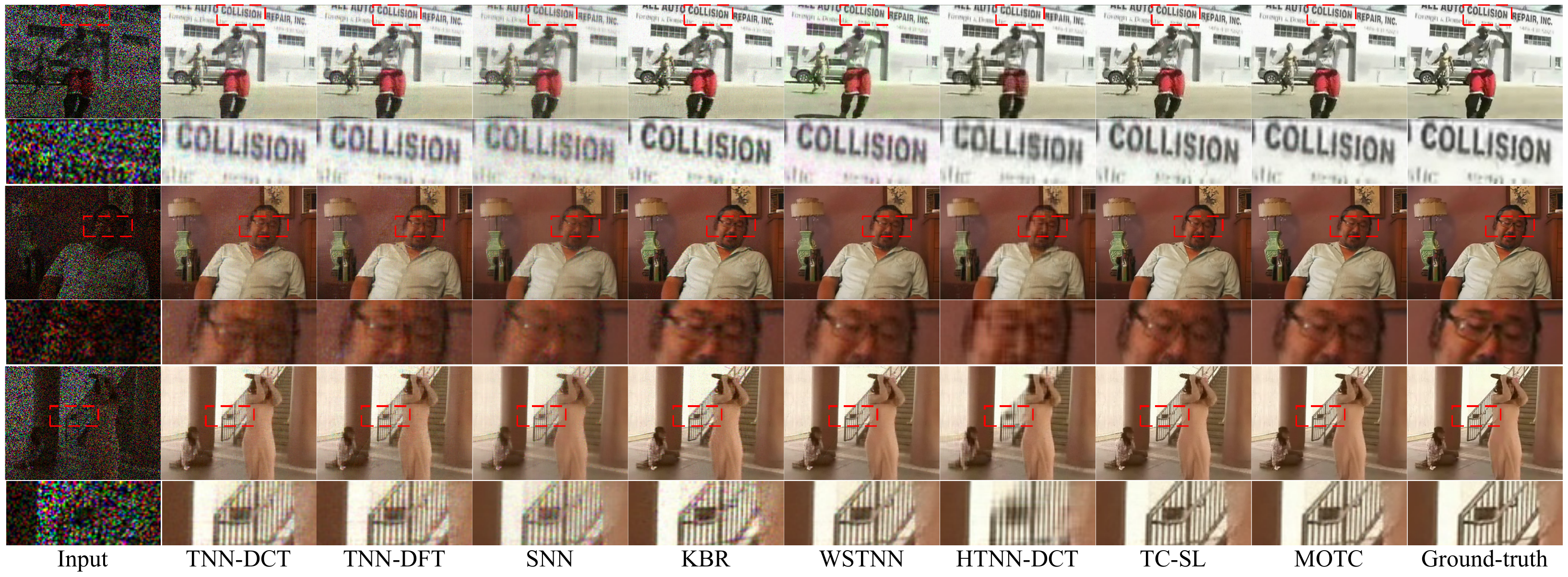} 
	\caption{Examples of video inpainting  by different methods on \textit{HMDB51} dataset for  case of $p=0.3$. \textbf{Best viewed in $\times 2$ sized color pdf file.} }\label{fig:fig_video_inpainting}
\end{figure}

\subsection{Color Video Inpainting for  Video with  Rapidly Changing Frames}\vspace{-0.15cm} 
 
We   evaluated all tensor completion methods on the randomly selected 50 color video segments with the   rapidly changing frames from the `run' category  of the \textit{HMDB51}\footnote{https://serre-lab.clps.brown.edu/resource/hmdb-a-large-human-motion-database/}.

We presented the PSNR values of all methods on the 50 video segments in Fig.~\ref{video_bar} and report their average results in Table \ref{videoe}. The results  show a significant improvement achieved by our methods (TC-SL and MOTC) for color video inpainting. For some videos (such as the 24-th, 34-th, 42-th, 43-th),  as shown in the Fig.~\ref{video_bar}, MOTC even achieved a 5-10 dB improvement in PSNR value!
Furthermore, the average results obtained by MOTC outperform the third-best method  by more than 3.5 dB on average, where the second-best method is TC-SL. The comparison between TC-SL and MOTC  demonstrates the effectiveness of the proposed multi-objective tensor recovery framework in  exploring the low-rankness of high-order tensor data across its various dimensions. Additionally, comparing tensor methods that consider the low-rankness of a tensor along with only one of its dimensions, TC-SL has achieved a 6.5 dB improvement!   This substantial improvement showcased by TC-SL in color video inpainting  provides strong evidence for its effectiveness in high-order tensor completion, particularly in scenarios involving non-smooth changes between tensor slices. Visual examples of video inpainting by different  methods are presented in Fig.~\ref{fig:fig_video_inpainting}. As we can see from the Fig.~\ref{fig:fig_video_inpainting}, even for cases that are hardly recognized by humans, MOTC can still reconstruct the video well and restore more detailed information than other tensor completion methods.



\section{Conclusions}   
In this work, we propose a multi-objective tensor recovery framework with a learnable tensor nuclear norm, which is solved by the proposed APMM-based heuristic optimization. This framework first provides an effective  solution for addressing the non-smooth   challenge  in  t-SVD-based methods. Thanks to our proposed methods, there has no longer a need to introduce $\tbinom{h}{2}$ variables and tune weighted parameters to analyze the correlation information of tensors across different dimensions. Experimental results in real-world applications demonstrate the superiority of our methods over previous methods, especially for the tensor data with non-smooth changes.

  It is worth noting  that the proposed tensor completion methods and framework  extend beyond tensor completion alone and are applicable to various tensor analysis problems and tasks, including data processing, representation learning, sketching, and clustering.  Many directions for future work are
possible.



\subsubsection*{Acknowledgements}
 We acknowledge  funding from New Frontiers in Research Fund (grant NFRFE-2022-00663), the NSERC Discovery Program (grants RGPIN-2019-05499 and   DGECR-2020-00296), the National Natural Science Foundation of China (grant U2033210), and  the Zhejiang Provincial Natural Science Foundation (grant LDT23F02024F02). We gratefully acknowledge the computing resources provided by Digital Research Alliance of Canada (\href{www.alliancecan.ca}{alliancecan.ca}).

%
%
\bibliographystyle{splncs04}
\bibliography{egbib}
\end{document}